\documentclass{article}

\usepackage[preprint]{neurips_2022}

\usepackage[utf8]{inputenc} 
\usepackage[T1]{fontenc}    
\usepackage{hyperref}       
\usepackage{url}            
\usepackage{booktabs}       
\usepackage{amsfonts}       
\usepackage{nicefrac}       
\usepackage{microtype}      
\usepackage{xcolor}         
\usepackage{graphicx}
\usepackage{xspace}
\usepackage{float}
\usepackage{amsmath}
\usepackage{subfig}

\title{Physics-informed Neural Network: The Effect of Reparameterization in Solving Differential Equations}

\author{%
  Siddharth Nand\\
  Department of Mathematics\\
  The University of British Columbia \\
  \texttt{sidnand@student.ubc.ca} \\
  \And
  Yuecheng Cai \\
  Department of Mechanical Engineering \\
  The University of British Columbia \\
  \texttt{ycai05@mail.ubc.ca}\\
}

\begin{document}

\maketitle

\begin{abstract}

	Differential equations are used to model and predict the behaviour of complex systems in a wide range of fields, and the ability to solve them is an important asset for understanding and predicting the behaviour of these systems. Complicated physics mostly involves difficult differential equations, which are hard to solve analytically. In recent years, physics-informed neural networks have been shown to perform very well in solving systems with various differential equations. The main ways to approximate differential equations are through penalty function and reparameterization. Most researchers use penalty functions rather than reparameterization due to the complexity of implementing reparameterization. In this study, we quantitatively compare physics-informed neural network models with and without reparameterization using the approximation error. The performance of reparameterization is demonstrated based on two benchmark mechanical engineering problems, a one-dimensional bar problem and a two-dimensional bending beam problem. Our results show that when dealing with complex differential equations, applying reparameterization results in a lower approximation error.
  
\end{abstract}

\section{Introduction}
Differential equations are important because they can be used to model a wide variety of phenomena that occur in the physical world, such as the movement of fluids, the behaviour of financial markets, the growth of populations, and the electrical current in a circuit. By solving a differential equation, we can find out how a system will behave over time and make predictions about its future behaviour. This can be useful in a wide range of fields, including engineering, physics, economics, and biology.

There exist two ways of solving differential equations, analytically and numerically. Analytical solutions give us an exact equation, but there are many differential equations whose exact solutions can't be found or are too hard to find. Numerical methods allow us to approximate a solution, but can't give us an exact formula. Recent developments in machine learning have led to neural networks (NNs), which are also universal approximators. This means a feedforward neural network with one hidden layer could accurately predict any continuous function arbitrarily well if there are enough hidden units  [1]. Therefore, we can use NNs to approximate the solution of a differential equation.

\subsection{Origins}

 The first paper outlining a method to approximate exact solutions for both ordinary differential equations (ODEs) and partial differential equations (PDEs) was published in 1997. In it, the authors converted a second-order differential equation (DE) into an optimization problem where the original function is approximated by adjusting the parameters of a neural network and using automatic differentiation to find the derivatives of the function [2]. 

\subsection{Physics-informed Neural Network}
 
After about two decades, Raissi et al. [3] introduced the term "physics-informed neural network" (PINN) and described a method for using deep neural networks to solve DEs. Since most natural phenomena and physics are governed by DEs, applying the PINN can significantly improve the accuracy of the model. By incorporating the laws of physics into the unsupervised learning process, PINN can decrease or even eliminate the need for training data. In many fields of research, obtaining training data is difficult due to expensive experimental or high-fidelity simulations; PINN can solve this problem. Benefiting from this, many researchers have proposed various techniques based on PINN in solving different problems, such as in fluid dynamics and electricity transmission [4].

\subsection{Current Research}

In the field of mechanical engineering, PINN shows great performance in predicting nonlinear behaviour of various structures. All these researchers applied PINN by adding penalty terms instead of reparameterization. For instance, by minimizing the potential energy of a structural system. Without any supervised learning phase, the trained unsupervised NN framework shows a high level of consistency compared to the analytical results in problems such as bending a beam in 2D, 3D geometry [5], and truss structures [6]. In addition to minimizing potential energy, the integrated physics knowledge can be engineered for constraints and boundary conditions, which has been demonstrated in modelling the bending plate [7], and 3D beam with different materials [8]. One can also incorporate this technique into other types of NNs such as recurrent neural networks (RNNs), where multiple long-short term memory networks (LSTMs) are utilized to sequentially learn different time-dependent features [9].

Other pieces of literature have investigated the effect of reparameterization. For instance, Zhu et al [10] compares the performance of different reparameterization schemes in PINNs and discusses the trade-offs between accuracy and computational efficiency. The authors present a series of numerical examples to demonstrate the effectiveness of different reparameterization schemes and provide guidelines for choosing the best scheme for a given problem. In addition, some scholars also present a detailed analysis of the effect of reparameterization on the accuracy and stability of PINN solutions such as Tran et al [11] and Zhang et al [12], who discuss the benefits and limitations of different reparameterization techniques and demonstrate its effectiveness through different examples.

\subsection{Problem and Contribution}

Comparing reparameterized and not reparameterized versions of a neural network to solve DEs is an important problem because the way that people embed physics (handle constraints) is through reparameterization and penalty functions. However, most works of literature only use one method or the other, as discussed in section 1.3, but never use both methods and compare the results for the DEs they are solving. The problem we will be solving is to see if reparameterization on a neural network can lower the approximation error compared to a benchmark case of not reparameterizing the network (using only penalty functions). Therefore, our contribution will be to show that a reparameterized version of a neural network reduces or does not reduce the approximation error when solving mechanical engineering DEs using neural networks. This contribution will give a greater inside into where reparameterized models perform better than non-reparameterized models and vice-versa. This will help researchers make smarter decisions as to which method can maximize their approximation accuracy.

\section{Methods and Background Information}

In this section, we introduce some background information and our methods for solving DEs to the readers. We start by explaining what are the penalty function and reparameterization methods. Then we explain the finite differences method (FDM) for computing a numerical solution to DEs. Finally, we talk about the PINN we will be using for our tests.

\subsection{Penalty Function and Reparameterization}
We will use a second-order differential equation as an example.

Given a differential equation of the form
\begin{equation}
    G(\vec{x}, \Psi(\vec{x}), \nabla \Psi(\vec{x}), \nabla^2 \Psi(\vec{x})) = 0, \vec{x} \in D
\end{equation}
with certain boundary conditions (BC), where $\vec{x} = (x_1, x_2, ..., x_n) \in \mathbb{R}^n$, $D \subset \mathbb{R}^n$ and $\Psi(\vec{x})$ is the function to be approximated [2].

Assume $F(\vec x)$ is the output of a neural network with $\vec x$ as an input. The prediction $F_i (x_i)$ for input $x_i$ of a neural network can be written as a function of the form

\begin{equation}
    F_i (x_i) = v^T (h_n \circ h_{n - 1} \circ \dots \circ h(Wx_i))
\end{equation}

Then we can easily find the derivatives using automatic differentiation. Next, we let $A(\vec x)$ satisfy the boundary conditions. If $\vec F(b_1) = \vec c_1, ..., \vec F(b_n) = \vec c_n$ are the boundary conditions, with $\vec b_i$, set to constants $\vec c_i$, then $A(\vec x)$ is of the form

\begin{equation}
    A(\vec x) = \left \Vert F(\vec b_1) - \vec c_1 \right \Vert + \left \Vert F(\vec b_2) - \vec c_2 \right \Vert + ... + \left \Vert F(\vec b_3) - \vec c_n \right \Vert
\end{equation}

Then we can treat $\Psi (\vec x)$ as the following loss function

\begin{equation}
    Loss = \textnormal{minimum }( A(\vec x) + \left \Vert G(\vec{x}, F(\vec{x}), \nabla F(\vec{x}), \nabla^2 F(\vec{x})) \right \Vert )
\end{equation}
This is called the penalty function method, where the BCs are handled by adding penalty terms with penalty coefficient $\lambda$. In short, the boundary conditions are not in-forced with this method, rather, the NN must minimize $A(\vec x)$ such that $F(\vec b_i)$ is as close to $\vec c_i$ as possible.

An alternative way to handle BC is by reparameterization, which explicitly forces the NN to satisfy the BCs by modifying the output representation of the NN. Instead of having $F(\vec{x})$ be the approximator of $\nabla \Psi(\vec{x})$, the reparameterized output becomes:

\begin{equation}
    K (\vec{x}) = B(x) F(\vec{x})
\end{equation}
where $B(x)$ is a function of $x$ so that the reparameterized function $K (\vec{x})$ can naturally satisfy the BCs:
\begin{equation}
    \left \Vert K(\vec b_1) - \vec c_1 \right \Vert = \left \Vert K(\vec b_2) - \vec c_2 \right \Vert = ... = \left \Vert K(\vec b_n) - \vec c_n \right \Vert = 0
\end{equation}

\subsection{Finite Differences (FDM)}
To obtain the higher-order derivatives for the numerical output, we use finite differences in our NN learning process. Since the problems considered (see Section 3) are static problems, only the spatial domain is discretized. Solving a set of equations based on the Taylor expansion and ignoring the high-order terms, the formula for calculating the 1st, 2nd and 4th-order derivative is shown here:

\begin{equation}
    \frac{dF}{dx} = \frac{F(x + h) - F(x-h)}{2h}
\end{equation}
\begin{equation}
    \frac{d^2F}{dx^2} = \frac{F(x + h) - 2F(x) + F(x-h)}{h^2}
\end{equation}
\begin{equation}
    \frac{d^4F}{dx^4} = \frac{F(x + 2h) - 4F(x+h) + 6F(x) - 4F(x-h) +F(x-2h)}{h^4}
\end{equation}

\subsection{Physics-informed Neural Network}
\begin{figure}[H]
    \centering
    \includegraphics[scale=0.5]{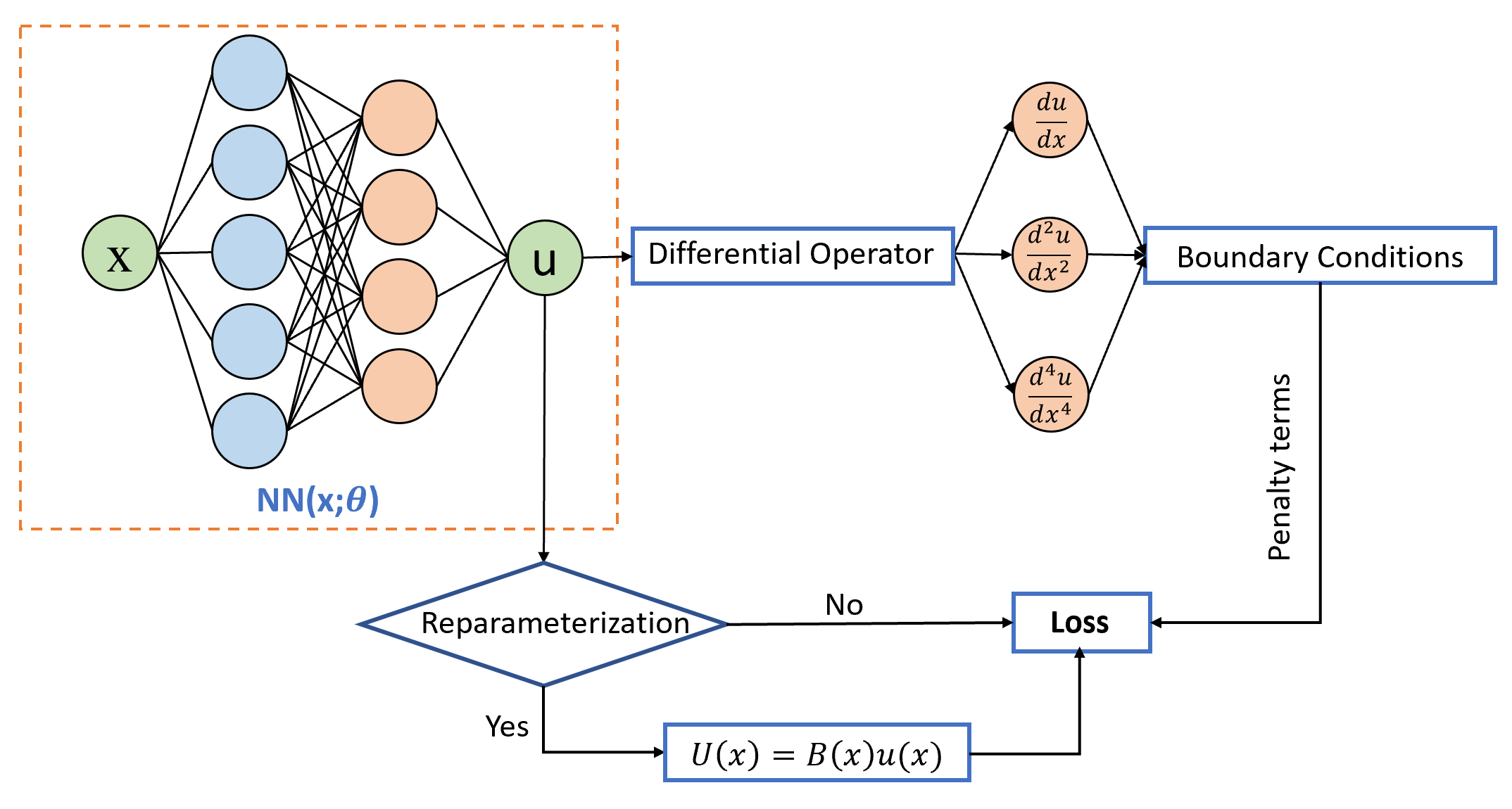}
    \caption{The general form of the PINN}
    \label{fig:NN}
\end{figure}

Fig. \ref{fig:NN} exhibits the general schematic of the PINN that we used to solve our DEs. The artificial neural network (ANN) is applied to approximate the mapping between $x$ and $u$ or $w$, depending on the purpose of the problem. A differential operator is employed to numerically calculate the derivatives of the output. The BCs are handled through penalty functions, reparameterization, or both.

In this project, the NN is composed of two hidden layers with 128 neurons in each layer. The training is performed using Adam optimization with an initial learning rate of 0.001. Sigmoid and ReLu activation functions are applied to the hidden and output layers, respectively. For the accuracy of the comparison, the same hyperparameter of the neural network is applied to all test cases.

\section{Problem Formulation}
In this section, we present two benchmark mechanical problems to demonstrate the effect of reparameterized and penalty functions. The implementation detail for both approaches of the two case studies will be exhibited.
\subsection{1D Bar problem}
\begin{figure}[H]
    \centering
    \includegraphics[scale=0.8]{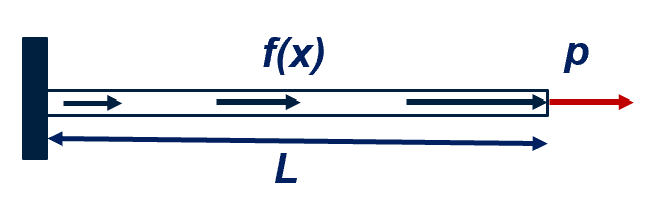}
    \caption{1D Bar}
    \label{fig:1d_bar}
\end{figure}

The first case study is a 1D bar problem, as described in Fig. \ref{fig:1d_bar}, The left side of the bar is fixed to the wall, while the right side is free. A non-uniformly distributed load $f(x)=x$ is applied through the central axis. An external concentrated force $P$ is employed at the right side of the bar. Assuming the length of the bar is $L$, the cross-sectional area is $A$ and the elastic Young's modulus is $E$ (a constant representing how easily a material can bend or stretch). The goal is to calculate the displacement of the bar along the x-axis.
Based on the beam theory, the governing equation for this problem can be represented as:\\
\begin{equation}
    -\frac{d}{dx}(EA\frac{du}{dx})=f(x)
\end{equation}

Where the boundary conditions are $u(x=0)=0$ and $u'(x=L)=0$, respectively. The first boundary condition states that there is zero displacement at the right side point, while the second boundary condition states that the velocity at the left side is zero.

\textbf{Case 1: 1D Bar Reparameterized}\\
To naturally satisfy both boundary conditions, the reparameterized NN estimator of the first problem can be represented as:\\
\begin{equation}
    U(x)=xe^{-x}NN(x)
    \text{; where NN(x) is the neural network}
\end{equation}\\
Instead of finding the mapping between $x$ and $u$, an additional term $xe^{-x}$ is applied so that that the reparameterized function $U(x)$ can have the following properties: $U(x=0)=0$ and $U'(x=L)=0$. According to this formulation, the NN loss function can be defined as:\\
\begin{equation}
    Loss(x) = \left \Vert \frac{d^2 U(x)}{dx^2} + \frac{f(x)}{EA} \right \Vert _2
\end{equation}\\
where $x$ is a vector of grid points uniformly sampled along the bar from $0$ to $L$.

\textbf{Case 2: 1D Bar Penalty Function}\\
As stated in section 2, a penalty function is the most widely applied technique to handle boundary conditions of differential equations. Since the first case study involves two BCs, two penalty terms will be included in the loss function. Note that the reparameterized term is not considered in this scenario, thus the output $U$ is calculated based on $NN(x)$:\\
\begin{equation}
    U(x)=NN(x)
\end{equation}\\
The loss function based on the penalty function technique can be represented as:

\begin{equation}
Loss(x) = \left \Vert \frac{d^2 NN(x)}{dx^2} + \frac{f(x)}{EA} \right \Vert _2 + \lambda_1 \left \Vert NN(x = 0) \right \Vert _2 + \lambda_2 \left \Vert \left. \frac{d NN(x)}{dx} \right |_{x = L} - P \right \Vert _2
\end{equation}

where $\lambda_1$ and $\lambda_2$ represent the penalty coefficients. They have been set to 100 for this project based on the preliminary investigations. 

\subsection{2D Bending Beam Problem}
\begin{figure}[H]
    \centering
    \includegraphics[scale=0.8]{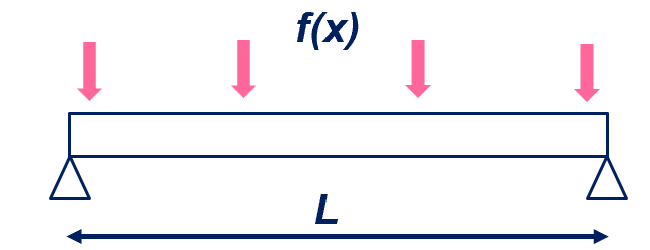}
    \caption{2D Bending Beam}
    \label{fig:2d_bar}
\end{figure}

The second case study is to estimate the vertical deflection (difference in direction between Earth's gravity and some reference direction) of a beam under non-uniformly distributed loading, as seen in Fig. \ref{fig:2d_bar}. Similar to the 1D bar case study, we assume the length of the beam is $L$, the Young's modulus is $E$, the second moment of inertia of the beam cross-section is $I$ (a measure of how resistant a cross-section is to bending), and the non-uniformly distributed loading as $f(x)=sin(x)$. Both sides of the beam are simply supported, so the beam is free to rotate, and the bending moment at both ends is zero. Based on the classical beam theory, the governing equation and the boundary conditions of this problem are:\\
\begin{equation}
    EI \frac{d^4w(x)}{dx^4}+f(x)=0 \label{eq:2d_bending_beam}
\end{equation}
Subject to:
\begin{equation}
    w(x=0)=0; w(x=L)=0; w''(x=0)=0; w''(x=L)=0 \label{eq:2d_bending_beam_constraints}
\end{equation}

\textbf{Case 3: 2D Beam Reparameterization}\\
In this 2D bending beam case, it is difficult to fully satisfy all the BCs using reparameterization, due to the existence of zero and the second-order derivative. Thus, part of the BCs will be handled through the penalty functions. The NN approximator after partially reparameterized can be represented as:
\begin{equation}
    W(x) = sin(\pi \frac{x}{L})NN(x)
\end{equation}
The corresponding loss function in this case is:
\begin{equation}
Loss(x) = \left \Vert \frac{d^4 W(x)}{dx^4} + \frac{f(x)}{EI} \right \Vert _2 + \lambda_1 \left \Vert \left. \frac{d^2 W(x)}{dx^2} \right |_{x = 0} \right \Vert _2 + \lambda_2 \left \Vert \left. \frac{d^2 W(x)}{dx^2} \right |_{x = L} \right \Vert _2 \label{eq:2d_beam_reparam}
\end{equation}
where $W(x)$ is the vertical deflection estimation. The first term in Eq. \ref{eq:2d_beam_reparam} shows the residual of the governing Eq. \ref{eq:2d_bending_beam}, and the last two terms represent the residual of the second-order BCs in Eq.\ref{eq:2d_bending_beam_constraints}.

\textbf{Case 4: 2D Beam Penalty Function}\\
Similar to case 1, no reparameterization will be considered in this case and the loss function is defined based on the residuals of governing equations and BCs. Therefore, the loss function of this 2D beam problem without reparameterized can be represented as:
\begin{equation}
Loss(x) = \left \Vert \frac{d^4 NN(x)}{dx^4} + \frac{f(x)}{EI} \right \Vert _2 + \lambda_1 \left \Vert \left. \frac{d^2 NN(x)}{dx^2} \right |_{x = 0,x = L} \right \Vert _2 + \lambda_2 \left \Vert \left. \frac{d^2 NN(x)}{dx^2} \right |_{x = 0,x = L} \right \Vert _2 \hspace{-.5em} \label{eq:2d_beam_penalty_func}
\end{equation}
where $NN(x)$ is the vertical deflection estimation and the last two terms in Eq. \ref{eq:2d_beam_penalty_func} include all the BCs in Eq. \ref{eq:2d_bending_beam_constraints}.

\section{Results}
According to the loss function we defined in Section 3, the NNs have been trained and the results are shown below. Fig. \ref{fig:case_1} and Fig. \ref{fig:case_2} represent the approximated displacement versus the analytical result at each node point for case 1 and case 2 respectively. It can be observed that the PINN shows good performance for both cases. The approximation error in cases 1 and 2 are 0.8063\% and 0.8136\% respectively. This indicates that the difference between the reparameterization and the penalty function method is not significant in the 1D bar case. Since only the first and second-order derivatives are included in the governing equations and BCs, this makes it easy for the neural network to estimate the target equations.

\begin{figure}[H]
    \hspace{-1.3in}
    \subfloat[1D Bar Reparameterization]{\includegraphics[width=0.8\textwidth]{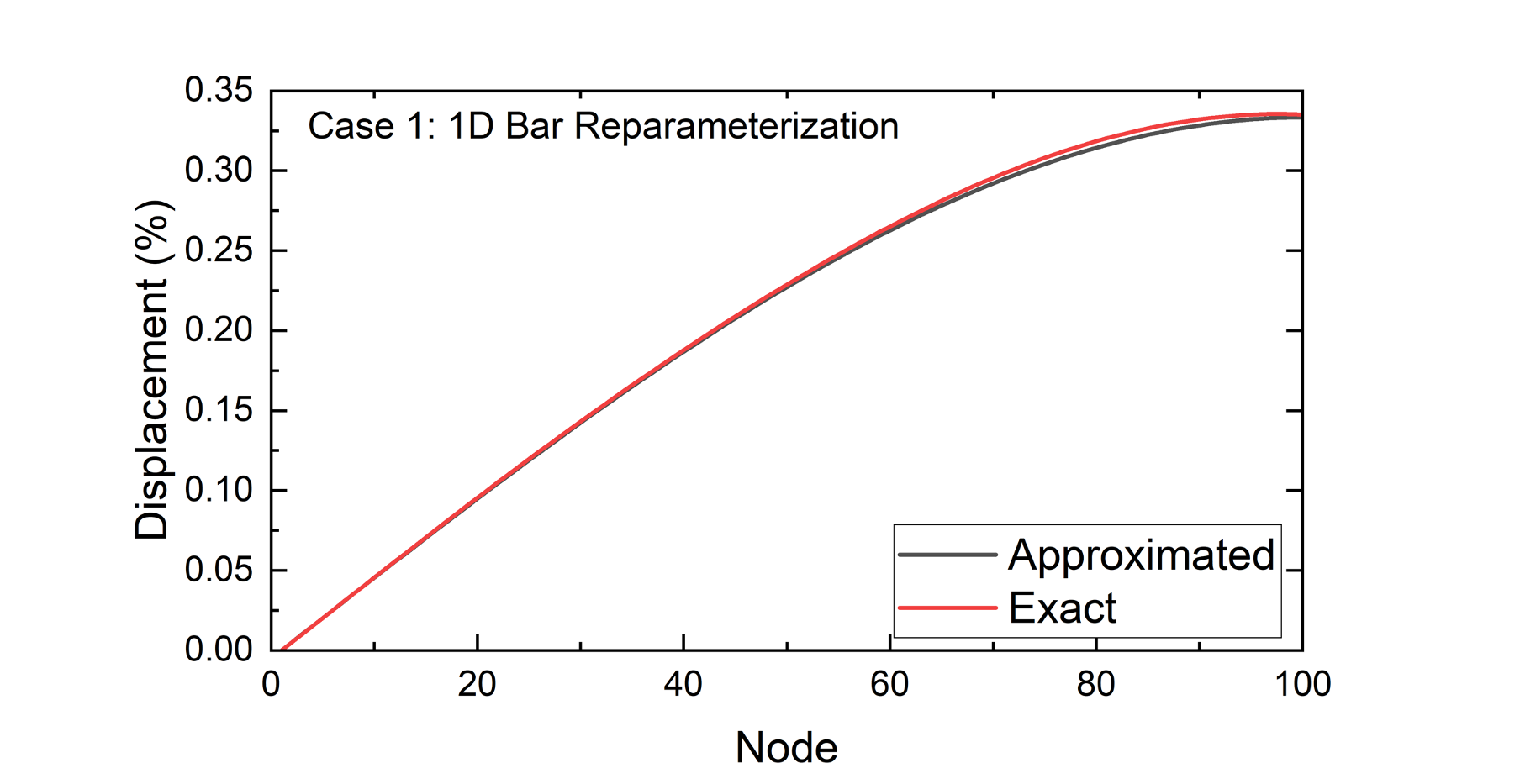}\label{fig:case_1}}
    \hspace{-0.5in}
    \subfloat[1D Bar Penalty Function]{\includegraphics[width=0.8\textwidth]{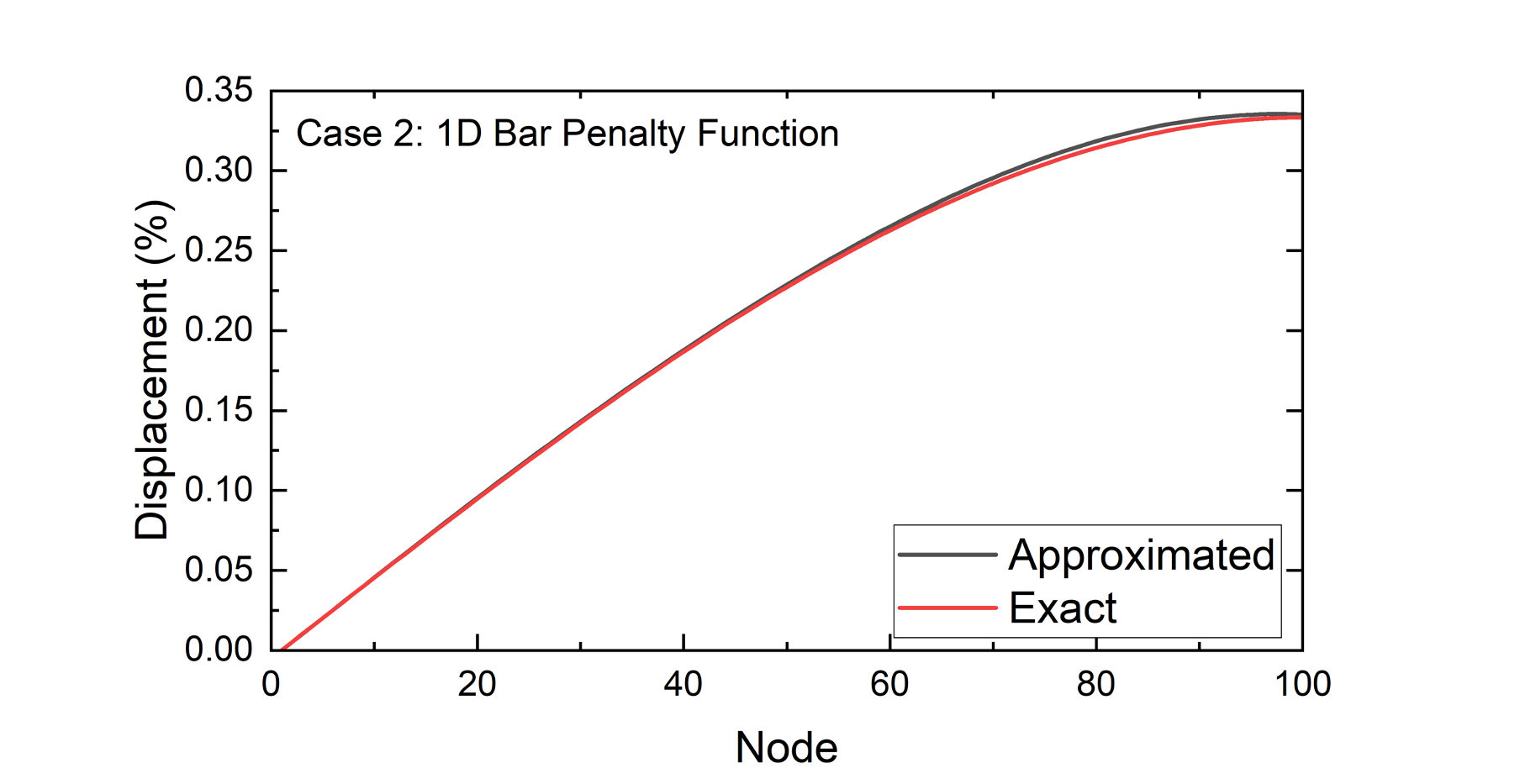}\label{fig:case_2}}
    \caption{1D Bar Test Results}
\end{figure}

\begin{figure}[H]
    \hspace{-1.3in}
    \subfloat[2D Bar Reparameterization]{\includegraphics[width=0.8\textwidth]{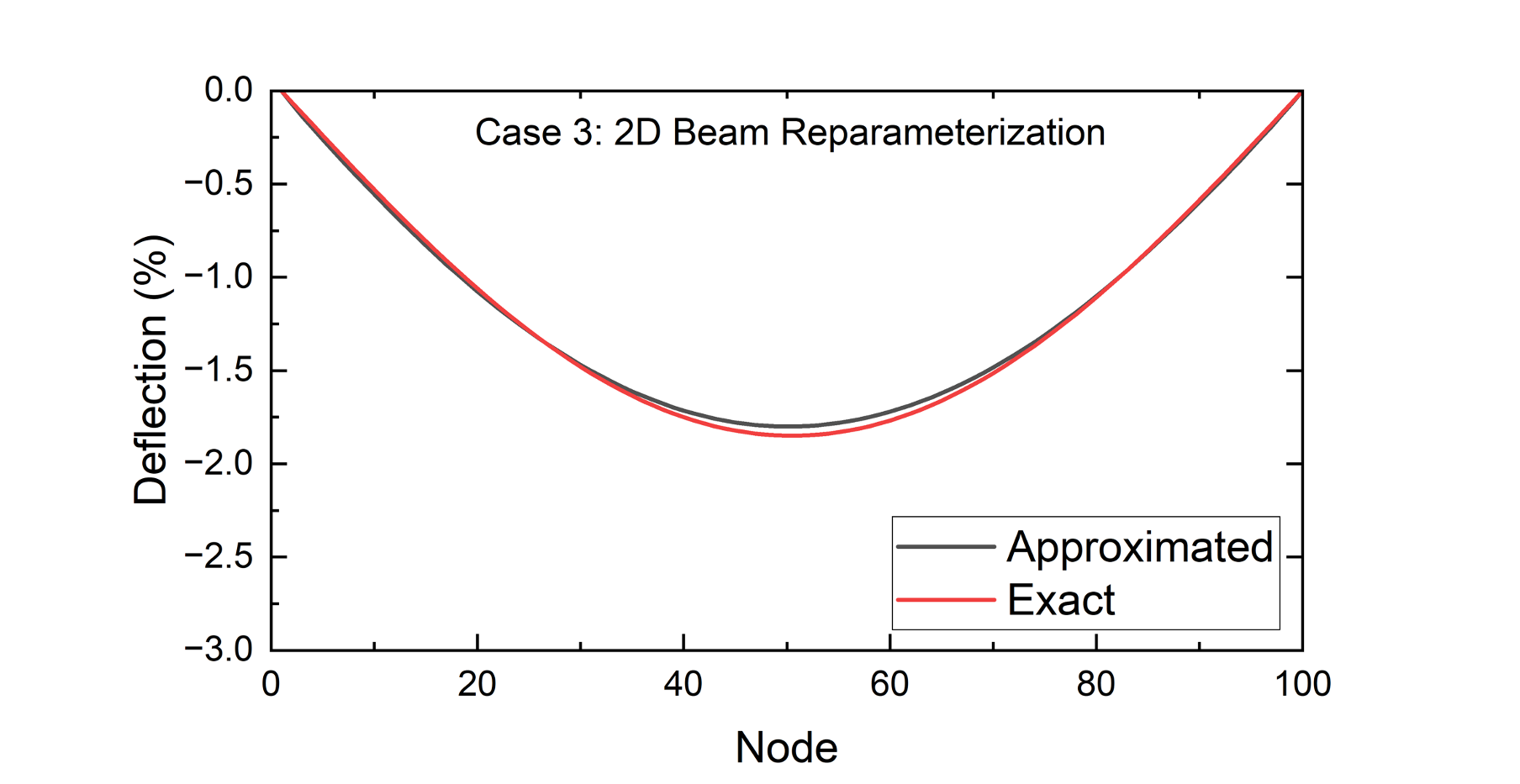}\label{fig:case_3}}
    \hspace{-0.5in}
    \subfloat[2D Bar Penalty Function]{\includegraphics[width=0.8\textwidth]{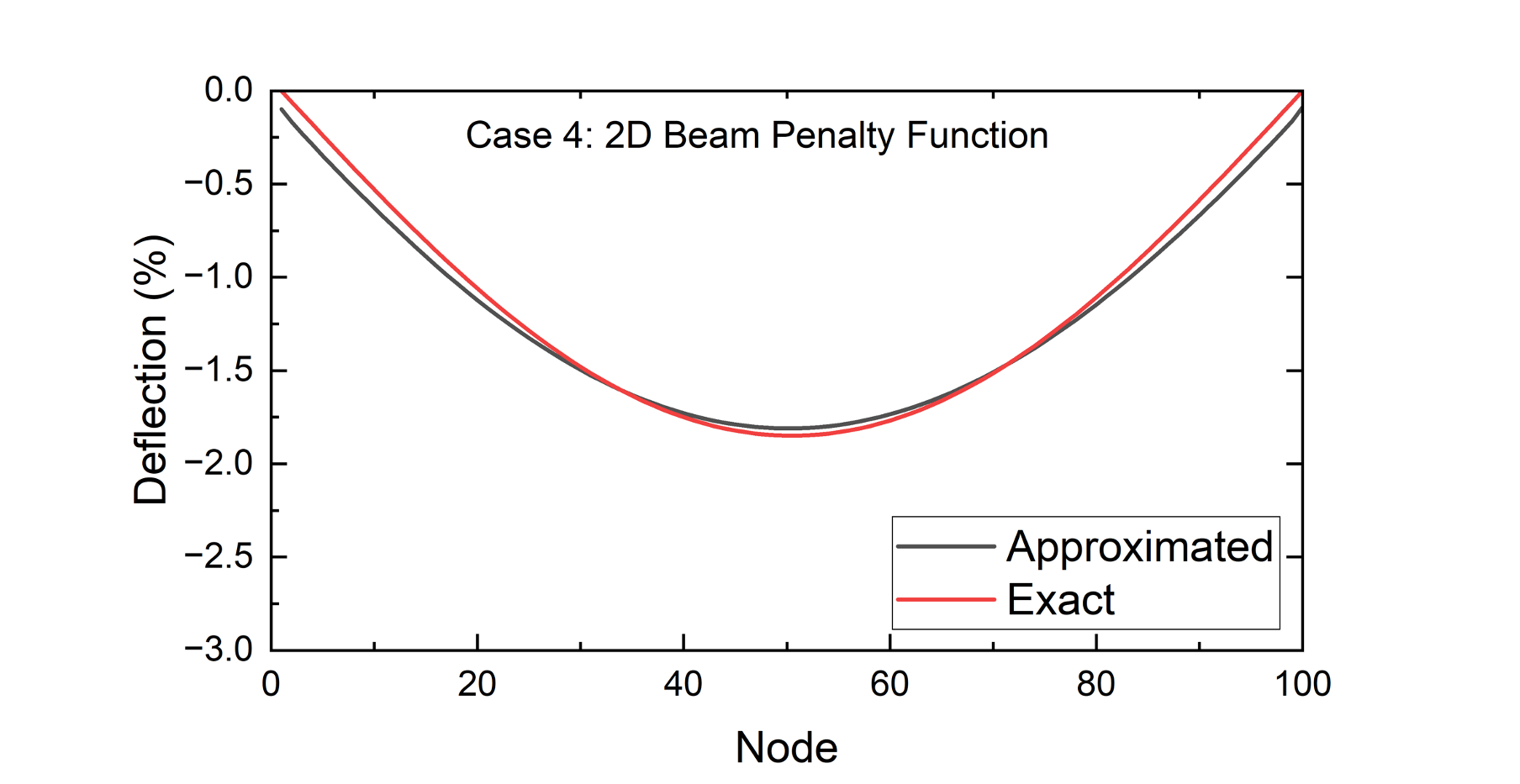}\label{fig:case_4}}
    \caption{2D Bar Test Results}
\end{figure}

However, when we remove the reparameterization in the 2D beam problem, the approximation noticeably deviates from the exact solution. Similar to Fig. \ref{fig:case_1} and Fig. \ref{fig:case_2}, Fig. \ref{fig:case_3} and Fig. \ref{fig:case_4} show that the approximated vertical deflection point is very close to the exact solution. It can be easily noticed that the result in case 3 shows zero deviation at both ends of the beam ($x=0$ and $x=L$). On the other hand, the penalty function does not manage to set the BCs to exactly zero and therefore, shows small deviations at both ends. This also affects the intermediate approximation of the results, at the first and last 30\% of the approximated curve in \ref{fig:case_4} show a large error. This is extremely important in real-life physics and engineering, where errors in initial conditions may be amplified by further steps. The approximation accuracy can be significantly improved even though the output is partially parameterized. The importance of reparameterization can also be observed from the approximation error in both cases, the errors are 2.7813\% and 12.9187\% for cases 3 and 4 respectively. In the case of the beam problem, the errors may come from the second-order BC and fourth-order governing equations. In this case, FDM may not be suitable for dealing with higher-order derivatives, which can be investigated in future studies.

\section{Conclusion}
In this paper, a physics-informed neural network is applied to solve two mechanical engineering benchmark problems involving different differential equations. The effect of reparameterization has been discussed based on the definition of the loss functions for both cases. The hyperparameter of the PINN for all test cases is the same. Based on the results in Section 4, we can conclude that:

1. Reparameterization shows a similar level of accuracy to the penalty function when the underlying DE is simple, evident in cases 1 and 2;

2. When dealing with difficult DEs, reparameterization dominates the penalty function in terms of approximation accuracy, evident in comparing cases 1 and 3 with 3 and 4;

3. Even partial reparameterization can significantly improve the PINN's overall performance, as evident in case 3.

A strength of our contribution is that we filled a research gap; no researchers have investigated the effect of reparameterization on the approximated result. Some weaknesses of our findings is the use of only ODEs and a lack of variety in our case studies. Our case studies only use ODEs, so our findings do not encompass PDEs. In addition, we only used two DEs, which is not a very large sample size of DEs. With such few DEs, our results are biased towards a small subset of DEs. Given more time, we would like to address our contribution's weakness of not testing PDEs and using a lot more DEs. This would provide more evidence for or against our findings and make our finds more reputable. Finally, future work can be placed on how to incorporate more accurate differential operators into PINN. In addition, more research needs to go into creating efficient ways to perform reparameterization according to different BCs because each NN needs to be reparameterized using a different function based on BCs, there is no systematic way to finding a function that will ensure the BCs are satisfied.

\section*{References}

{
\small
    [1] Hornik, K., Stinchcombe, M., \& White, H. (1989). Multilayer feedforward networks are universal approximators. Neural Networks, 2(5), 359–366. https://doi.org/10.1016/0893-6080(89)90020-8

    [2] Lagaris, I. E., Likas, A., \& Fotiadis, D. I. (1997). Artificial neural networks for solving ordinary and partial differential equations. Retrieved from http://arxiv.org/abs/physics/9705023

    [3] Raissi, M., Perdikaris, P., \& Karniadakis, G. E. (2017). Physics informed deep learning (part I): Data-driven solutions of nonlinear partial differential equations. Retrieved from http://arxiv.org/abs/1711.10561

    [4] Karniadakis, G. E., Kevrekidis, I. G., Lu, L., Perdikaris, P., Wang, S., \& Yang, L. (2021). Physics-informed machine learning. Nature Reviews Physics, 3(6), 422–440. https://doi.org/10.1038/s42254-021-00314-5

    [5]	V. M. Nguyen-Thanh, X. Zhuang, and T. Rabczuk, “A deep energy method for finite deformation hyperelasticity,” Eur. J. Mech. - ASolids, vol. 80, p. 103874, Mar. 2020, doi: 10.1016/j.euromechsol.2019.103874.
    
    [6]	H. T. Mai, Q. X. Lieu, J. Kang, and J. Lee, “A robust unsupervised neural network framework for geometrically nonlinear analysis of inelastic truss structures,” Appl. Math. Model., vol. 107, pp. 332–352, Jul. 2022, doi: 10.1016/j.apm.2022.02.036.
    
    [7]	H. Guo, X. Zhuang, and T. Rabczuk, “A Deep Collocation Method for the Bending 
    Analysis of Kirchhoff Plate,” Comput. Mater. Contin., vol. 59, no. 2, pp. 433–456, 2019, doi: 10.32604/cmc.2019.06660.
    
    [8]	D. W. Abueidda, Q. Lu, and S. Koric, “Meshless physics‐informed deep learning method for three‐dimensional solid mechanics,” Int. J. Numer. Methods Eng., vol. 122, no. 23, pp. 7182–7201, Dec. 2021, doi: 10.1002/nme.6828.
    
    [9]	R. Zhang, Y. Liu, and H. Sun, “Physics-informed multi-LSTM networks for metamodeling of nonlinear structures,” Comput. Methods Appl. Mech. Eng., vol. 369, p. 113226, Sep. 2020, doi: 10.1016/j.cma.2020.113226.

    [10] Zhu, W., Xu, K., Darve, E., Biondi, B., \& Beroza, G. C. (2021). "Integrating deep neural networks with full-waveform inversion: Reparameterization, regularization, and uncertainty quantification". GEOPHYSICS, 87(1), R93–R109. https://doi.org/10.1190/geo2020-0933.1

    [11] Tran T , Razavi A., and Bazant M. Z. (2021) "Reparameterization in Physics-Informed Neural Networks for Improved Accuracy and Stability"

    [12] Zhang J., Fan Y., and Ying L. (2020) "Reparameterization in Physics-Informed Neural Networks: A Comparative Study"
}

\end{document}